\documentclass[runningheads]{llncs}
\usepackage{graphicx}
\usepackage{amsmath,amssymb} 
\usepackage{color}

\begin{document}

\newcommand{\point}{
    \raise0.7ex\hbox{.}
    }


\pagestyle{headings}

\mainmatter

\title{Automatic Visual Theme Discovery from Joint Image and Text Corpora} 



\author{Ke Sun, Xianxu Hou, Qian Zhang, Guoping Qiu} 
\institute{School of Computer Science, University of Nottingham Ningbo China \\
	(Ke.Sun, Xianxu.Hou, Qian.Zhang, Guoping.Qiu)@nottingham.edu.cn} 

\maketitle

\begin{abstract}
A popular approach to semantic image understanding is to manually tag images with keywords and then learn a mapping from visual features to keywords. Manually tagging images is a subjective process and the same or very similar visual contents are often tagged with different keywords. Furthermore, not all tags have the same descriptive power for visual contents and large vocabulary available from natural language could result in a very diverse set of keywords. In this paper, we propose an unsupervised visual theme discovery framework as a better (more compact, efficient and effective) alternative to semantic representation of visual contents. We first show that tag based annotation lacks consistency and compactness for describing visually similar contents. We then learn the visual similarity between tags based on the visual features of the images containing the tags. At the same time, we use a natural language processing technique (word embedding) to measure the semantic similarity between tags. Finally, we cluster tags into visual themes based on their visual similarity and semantic similarity measures using a spectral clustering algorithm. We conduct user studies to evaluate the effectiveness and rationality of the visual themes discovered by our unsupervised algorithm and obtains promising result. We then design three common computer vision tasks, example based image search, keyword based image search and image labelling to explore potential application of our visual themes discovery framework. In experiments, visual themes significantly outperforms tags on semantic image understanding and achieve state-of-art performance in all three tasks. This again demonstrate the effectiveness and versatility of proposed framework.
\end{abstract}

\section{Introduction}
The popularisation of photo sharing websites such as Flickr and Instagram encourages more and more people to share their life experience by uploading numerous images on a daily basis. Among various information contained in these images, associated tags are of great importance for helping computer vision (CV) algorithms to understand the semantic meaning of images. However, due to human's subjectivity towards visual content understanding, different tags are often used to describe visually similar images. For example, images containing lakes, rivers and ocean could all be tagged with \textit{water}, or we can just use \textit{lake}, \textit{river}, \textit{ocean} respectively. It's a very natural case but it often confuses CV algorithms since they are forced to distinguish similar visual instances. 

Additionally, not all tags show strong connection to particular visual content, such as \textit{wonderful} and \textit{beautiful}, but they are frequently used in social media websites. Visually describing such kinds of tags is pretty challenging even for human themselves, let alone CV algorithms. 

Another problem is the curse of dimensionality. Appearance of different tags in image annotations are often represented using one-hot encoding in order to be easily processed by CV algorithms. Hence, the dimensionality of annotation for a single image is the whole size of tag vocabulary. The overall annotation matrix would be extremely sparse since each image is only associated with a few tags. Traditional dimensionality reduction methods mainly focus on tag frequency, while the semantic and visual correlation between tags are often ignored. 

\begin{figure*}[th]
	\centering
	\centerline{\includegraphics[width=12cm,height=5.5cm]{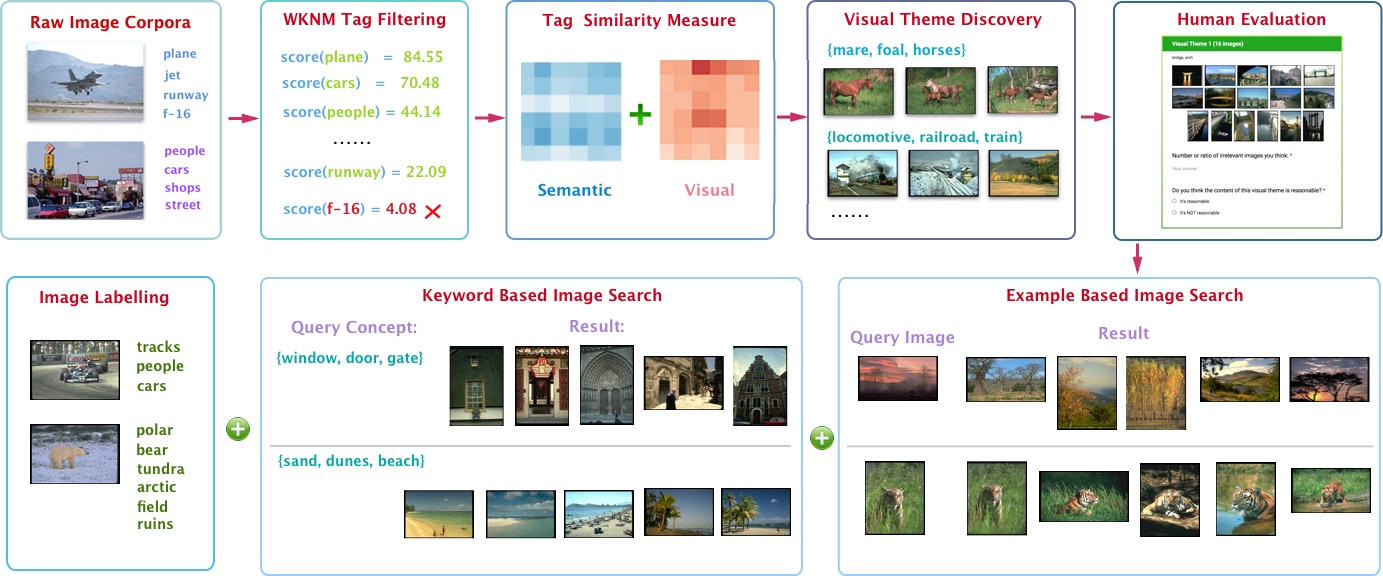}}
	\caption{Overview of visual theme discovery framework and its applications. Given images and associated tags, we first eliminate less qualified tags using WNKM tag filtering method, then clustering tag into visual themes according to their semantic and visual similarities. Next we ask human evaluators to evaluate the quality of discovered visual themes. Applications of visual themes are shown at the bottom row.}
	\label{fig:pipeline}
\end{figure*}

To address the issues mentioned above, We propose to use Visual Theme (VT) as a replacement of tag-based annotation for compact visual content description. A visual theme, consisted of a small set of tags, is capable of describing a group of similar visual contents in images. Besides, tags within the same VT are also semantically related.

We develop a data-driven framework to automatically discover VTs from joint image and tag corpora. We start by examining each tag's ability for visual content description, then eliminate tags whose descriptive ability fall under certain level. Next we measure the pairwise semantic and visual similarity amongst the remaining tags, then merge them into a joint similarity matrix. Visual similarity measures how tags are visually connected to describe visual contents, and semantic similarity measures how close tags are in natural language understanding. Finally we cluster tags into a collection of VTs according to the joint similarity matrix. The workflow of the proposed framework is illustrated in Fig.~\ref{fig:pipeline}.

In order to evaluate the quality of visual themes, we ask human evaluators to examine how well these themes are in the task of describing similar visual contents. The result is pretty promising and demonstrate the effectiveness and rationality of discovered visual themes. We also explore potential applications of discovered VTs by designing three common CV tasks: example based image search, keyword based image search and image labelling. We work on four popular benchmarks, namely, Corel5K \cite{duygulu2002object}, NUS-Wide-Lite \cite{chua2009nus}, IAPR-TC12 \cite{escalante2010the} and a subset of ESP-game \cite{guillaumin2009tagprop:}. The first two are used for example based search and keyword based search respectively. The last two and Corel5K are chosen as the testbeds of image labelling. We show the usefulness and advantages of using VTs rather than individual tags for these tasks.

\section{Related Work}
Our definition of visual theme is partly inspired by the naming of visual concept \cite{sun2015automatic}. A visual concept is denoted as a subset of human language vocabularies that refer to particular visual entities (e.g. fireman, policeman). Visual concepts have long been collected and used by computer vision researchers in multiple domains \cite{sadanand2012action}\cite{zhou2015conceptlearner}\cite{fang2015captions}\cite{singh2015selecting}. A example in image analysis is ImageNet \cite{deng2009imagenet}, where visual concepts (only nouns) are selected and organised hierarchically on the basis of WordNet \cite{miller1995wordnet}. A drawback of visual concepts is, they are often manually defined, and sometimes they may fail to capture complex information within the visual world. This makes them less applicable in multiple domains.

The subjectivity of visual concept definition hinders its extension to be used on different joint image and text databases. This motivates us to explore objective visual theme directly from raw images and associated tags. Our work on visual theme discovery is related to previous work on concept discovery \cite{sun2015automatic}\cite{sadeghi2015viske}\cite{divvala2014learning}. In particular, LEVAN \cite{divvala2014learning} starts with a given groups of general concepts and gradually divide them into subconcepts according to massive resources of online books. VisKE \cite{sadeghi2015viske} focuses on validating relationship between pairs of concept entities from semantic and visual aspects. \cite{sun2015automatic} builds a large amount of classifiers for terms filtering and similarity computation, then cluster selected terms into concepts. 

A significant difference between our work and previous work is that we are not trying to build large amount of general visual concepts so as to describe as many image as possible, instead, we put forward an unsupervised and efficient framework to allow different image databases to have their own collection of visual themes as visual content description. Considering the quantity and diversity of images, dividing large image collections into visual theme based categories can facilitate various tasks such as management, indexing and retrieval.

\section{Visual Theme Discovery}
This section elaborates the theme discovery workflow. Recall that a visual theme is constructed by a subset of tags which are capable of representing similar visual contents. To make it practical, we argue a VT should show strong connection to certain visual content that can be easily processed by computer vision algorithms. Besides, tags (including synonyms) describing same or similar visual content should be grouped into the same theme in order to maintain compactness. Start with the image corpus and associated tags, we first pick tags which show high-level visual content descriptive power, then cluster them into a set of VTs based on visual and semantic similarity.

\begin{figure}[ht]
	\centering
	\centerline{\includegraphics[width=10cm,height=4.5cm]{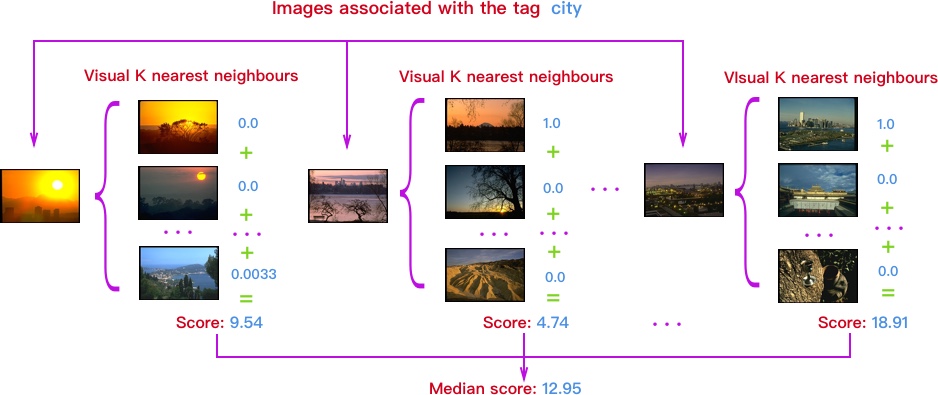}}
	\caption{Workflow of Weight K-Nearest Measure. Given a tag and its associated images, for each image, we find its visual K-nearest neighbours and examine if other images under the same tag frequently appear in the K neighbours. We compute a score (higher is better) of each associated image of given tag, then take the median to quantify the tag's ability towards visual content description.}
	\label{fig:wknm}
\end{figure}

\subsection{Tag Filtering}
As we mentioned before, not all tags show strong connection with visual contents. Before  discovering the themes, we need to examine each tag's ability of visual content description, and filter out ones who are not qualified. The idea to achieve this is simple: if a tag is good at describing particular visual content, the majority of its associated images should also share similar visual contents. Hence, the visual similarities between images under a tag can reflect the tag's ability towards visual content description. As for implementation, we represent images using feature activations from the pre-trained convolution neural network (CNN) model due to its excellent performance in content-based image retrieval \cite{babenko2014neural}. We then define Weighted K-Nearest Measure (WKNM) as measurement of tag's ability towards visual content description. 

The procedure of WKNM is illustrated in Fig. \ref{fig:wknm}. Given tag $t_{i}$ and its associated image set $F_{i} = set\left \{f_{i1}, f_{i2},...,f_{ij},...,f_{in} \right \}$, for every related image $f_{ij}$, K nearest neighbours based on the cosine distance of their visual features are obtained using cosine distance. Thus, the similarity score between image $f_{ij}$ and other images in $F_{i}$ could be computed as:

\begin{equation}
Sim(f_{ij}, F_{i}) = \sum_{k=1}^{K}(1-\frac{k-1}{K})\delta (t_{i}, f_{ijk})
\label{equ:wknm}
\end{equation}
where $\delta (t_{i}, f_{ijk})$ is an indicator function which equals to 1 if image $f_{ijk}$ contains tag $t_{i}$, otherwise it is set to 0. $K$ is the number of nearest neighbours of image $f_{ij}$. It could be noticed that, $\delta (t_{i}, f_{ijk})$ is penalised by multiplying a weight according to the sequence in K neighbours (a closer neighbour has a smaller sequence index). Hence, $Sim(f_{ij}, F_{i})$ quantifies tag $t_{i}$'s ability towards visual content description based on image $f_{ij}$.

We successively compute all similarity scores based on each image in tag $t_{i}$'s associated image set $F_{i}$, then take the median score to quantify tag $t_{i}$'s ability towards visual content description. We call such a median the Visual Content Descriptive Level (VCDL) of a given tag. A larger VCDL of a tag indicates it is good at describing certain visual contents. We choose the median because it is a robust statistic, even if dataset is biased, the median is unlikely to offer an arbitrarily large or small result. We repeat this procedure on all tags and eliminate those whose VCDLs fall below a certain threshold. Note that we do not need to examine each tag's frequency of occurrence since the WKNM method has inherently done this.

\begin{table}[ht]
	\begin{center}  
		\caption{Example of filtered tags on Corel5K dataset.}
		\begin{tabular}{l|c}  
			\hline
			Filtered tags & Evaluation\\ 
			\hline  
			\{f-16, kauai, oahu\} & too specific \\         
			\{whited-tailed, close-up\} & too abstract \\        
			\{art, festival\} & too generic\\ \hline
		\end{tabular}
	\end{center}
	\label{tab:tf}
\end{table}

Table 1 gives a few examples of filtered tags on Corel5K dataset. We could clearly see our method is able to automatically remove tags that are not suitable for visual theme discovery. However, when we take a look at the remaining tags, we found some of them are synonyms e.g. \textit{jet} and \textit{plane}. It is necessary to group them together since they are likely to confuse CV algorithms and introduce extra computational cost. Moreover, we notice some tags are often used together to describe particular visual content.  For instance, in Corel5K dataset, \textit{grizzly} only appears together with \textit{bears} in images containing bears. This motivates us to measure tag similarity both semantically and visually.

\subsection{Tag Visual Similarity Measure}
We measure tag visual similarity by examining their distance in the metric space. Suppose we have a visual space constructed by all images, and each image's distance could be evaluated by computing distances between their corresponding visual features.  In this space, each tag could be represented by its associated images which are a subset of the whole image set in visual space. Hence, the well-known Hausdorff distance (HD) is quite appropriate to measure visual distance between two different tags.

The Hausdorff distance is defined as the maximum distance of a set to the nearest point in the other set \cite{shonkwiler1991computing}. In our case, the Hausdorff distance from tag $A$ to tag $B$ in visual space would be:

\begin{equation}
h(A, B) = {\max_{a \epsilon A}} \{ { \min_{b \epsilon B}}  \{dist(a, b)\} \} 
\label{equ:hd}	
\end{equation}
where $a$ and $b$ are image feature based points of tags $A$ and $B$ in high-dimensional visual space, $dist(a, b)$ is certain distance metric between these points. For simplicity, we take $dist(a, b)$ as the Euclidian distance between $a$ and $b$. 

Since HD measures the relative position of points in visual space, it's more robust to position variations than other methods. However, HD method is quite sensitive to outliers, which makes it inappropriate to tackle noisy data. A modified version of HD is proposed in \cite{dubuisson1994modified}:

\begin{equation}
h_{mod}(A, B) =  \frac{1}{\left | A \right |} \sum_{a \epsilon A} {\min_{b \epsilon B} }  \{dist(a, b)\} 
\label{equ:hdmodi}	
\end{equation}
where ${\left | A \right |}$ is the number of images associated with tag $A$. A problem of this revised HD is it contains points whose pairwise distances are zero. Considering that an image's annotation often contains more than 1 tag, points $a$ and $b$ in $dist(a, b)$ could refer to the same image. Hence, we revise the formula in (\ref{equ:hdmodi}) to remove this negative impact:

\begin{equation}
h_{mod}^{'}(A, B) =  \frac{1}{\left | A^{'} \right |} \sum_{a \epsilon A} {\min_{b \epsilon B} }  \{dist(a, b)\} 
\label{equ:hdmodi2}
\end{equation}
where $A^{'}= {\min_{ b \epsilon B}} \{ dist(a, b) \neq 0 \}$. We use (\ref{equ:hdmodi2}) to measure visual distance from tag $A$ and tag $B$. Since associated images of different tags also differ, we modify the final Hausdorff distance between two tags as:

\begin{equation}
F^{'}(A, B) = \max \{h_{mod}^{'}(A, B), h_{mod}^{'}(B, A) \}
\label{equ:hdfinal}
\end{equation}

Ultimately we can obtain a distance matrix $M_{vdist}$ where each entry is the visual distance between two tags. It's easy to switch distance to similarity: just rescale all values in $M_{vdist}$ to the range from 0 to 1, then replace each entry value with the difference between 1 and original value. We denoted the tag visual similarity matrix as $M_{vsim}$. Larger values in $M_{vsim}$ indicates stronger visual similarity between two corresponding tags.

\subsection{Tag Semantic Similarity Measure}
We measure semantic similarity between two tags by evaluating their word embeddings \cite{mikolov2013distributed} \cite{bengio2006neural} in an unsupervised manner. In the embedding space, each distinct word is represented using a N-dimensional vector. The embedding algorithm first assign each word vector with random values, then recursively adjust the value of these vectors according to some objective function. More specifically, we train a Skip-gram neural network language model \cite{mikolov2013distributed} on latest dump of English Wikipedia using Word2Vec \cite{mikolov2014word2vec} toolset. 

To elaborate, the training set is a large collection of English Wikipedia articles. In the training phase, each time a short sequence of words are extracted from an article using a sliding window with fixed width. Then the corresponding word vectors (random values at first) are extracted and fed into the skip-gram model. The training objective is to enable words to effectively predict nearby words, so words enjoy higher semantic similarity lie closer in the semantic space. 

Once training process is completed, we extract word vectors from the trained model  according to the content of tags, then evaluate semantic similarity of each pair of tags by computing cosine distance between their corresponding word vectors. Similarly, we build the the semantic similarity matrix $M_{ssim}$. Again, we replace each entry value in $M_{ssim}$ with the difference between 1 and original value. Larger values in $M_{ssim}$ indicates stronger semantic similarity between two corresponding tags.

\subsection{Clustering Tags into Visual Themes}
With two similarity matrices $M_{vsim}$ and $M_{ssim}$, we linearly merge them into joint similarity matrix $M_{join}$ via a parameter $\alpha$ (from 0 to 1). We can control the proportion of visual and semantic components by tuning $\alpha$.

\begin{equation}
M_{join} = \alpha \times M_{vsim} + (1-\alpha) \times M_{ssim}
\end{equation}

Based on $M_{join}$, we use spectral clustering \cite{von2007tutorial} to cluster tags into a collection of visual themes. Table \ref{tab:vt} describes a few themes discovered on Corel5K dataset with $\alpha$ fixed to 0.12. Note that although $\alpha$ could be set to 0, which means no visual clue is used for similarity measurement, however, that might lead to sub-optimal result since semantic similarity mainly depends on word co-occurrence in text corpus.

\begin{table}[ht]
	\caption{Example of visual themes discovered on Corel5K dataset.}
	\begin{center}
		\begin{tabular}{l|l}  
			\hline
			Concept Type & Concept Content \\ \hline  
			scene & \{sunrise, sunset\} \\         
			object & \{mare, foals, horses\} \\        
			mixed & \{cloud, sky, mist, horizon\} \\
			mixed & \{jet, flight, runway, plane\} \\ \hline
		\end{tabular}
	\end{center}
	\label{tab:vt}
\end{table}

\section{Human Evaluation of Visual Themes}
After clustering phase, each visual theme is represented as a set of tags and associated images. As we mentioned in Section 3, a visual theme should show strong connection to certain visual content. Besides, tags (including synonyms) describing the same or similar visual content should be grouped into the same theme. Hence, we design a human evaluation experiment to examine quality of discovered visual themes from these two aspects.

\textbf{Evaluation setup}: We use Corel5K dataset and discover 100 visual themes using 4500 training images and associated tags. We feed training images into the VGG-16 \cite{simonyan2014very} model and take the output of fully-connected layer 'fc7' (4,096 dimensions) as image-level holistic visual features. Then we choose 499 testing images as evaluation set, and replace tag based annotation with corresponding visual themes. Hence, testing images are categorised into a collection of visual themes. Next we remove themes whose frequencies of occurrence are less than 3 times across all testing images, and keep 66 visual themes for evaluation. A print version of evaluation examples and interface could be found in the supplementary materials.

\begin{figure}[ht]
	\centering
	\centerline{\includegraphics[width=6cm,height=8cm]{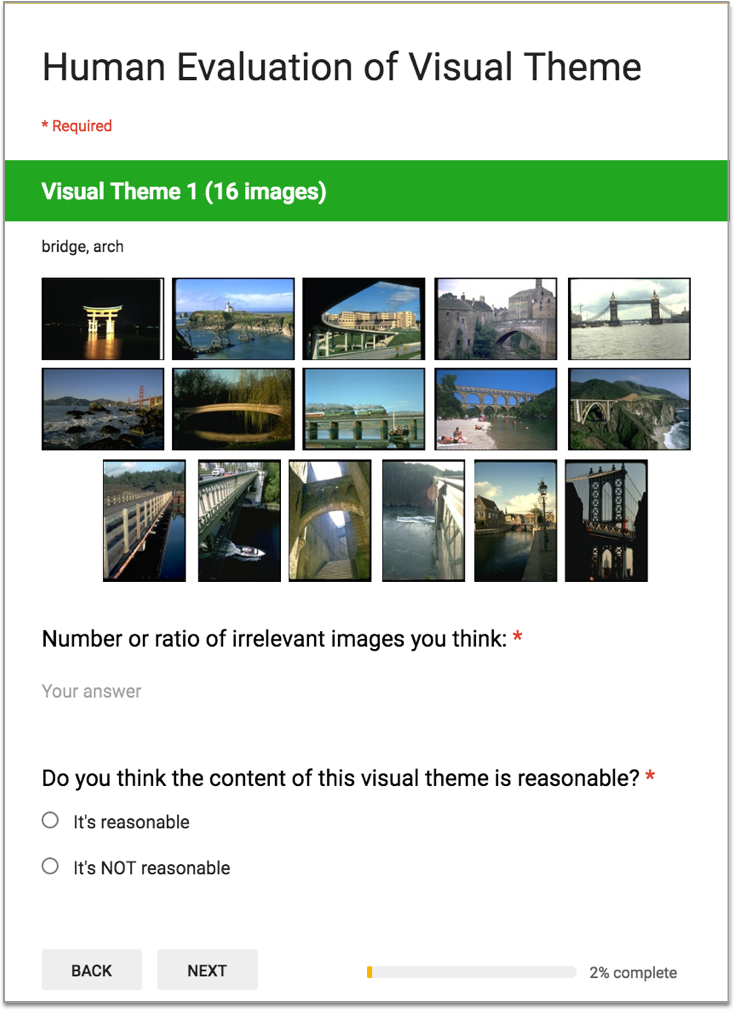}}
	\caption{An example of human evaluation interface. Human evaluators need to give numbers of images which are irrelevant to current displayed visual theme, and also vote for the rationality of this visual theme.}
	\label{fig:human_inter}
\end{figure}

We designed a two-step procedure for human evaluation. A example of the evaluation interface is shown in Fig. \ref{fig:human_inter}. For each visual theme, we first display its tags and associated images to human evaluators, then asked them to examine whether the visual content described by this visual theme appears in every associated images. If not, they need to give number of images which they think are relevant to the given theme. Thus we can easily compute the ratio of relevant images for each visual theme, and we name such a ratio as accuracy of visual content description (AVCD) of a visual theme. The AVCD for each visual theme is obtained by averaging all evaluators' responses on that theme. 

In the following step we asked evaluators to examine tags contained in visual themes. They need to check if all tags within a visual theme are semantically connected and refer to similar visual content. If so, the corresponding visual theme is regarded as rational and vice versa. The final decision of rationality for each visual theme was combined using majority vote of human evaluators.

\begin{figure}[ht]
	\centering
	\centerline{\includegraphics[width=9.5cm,height=4.8cm]{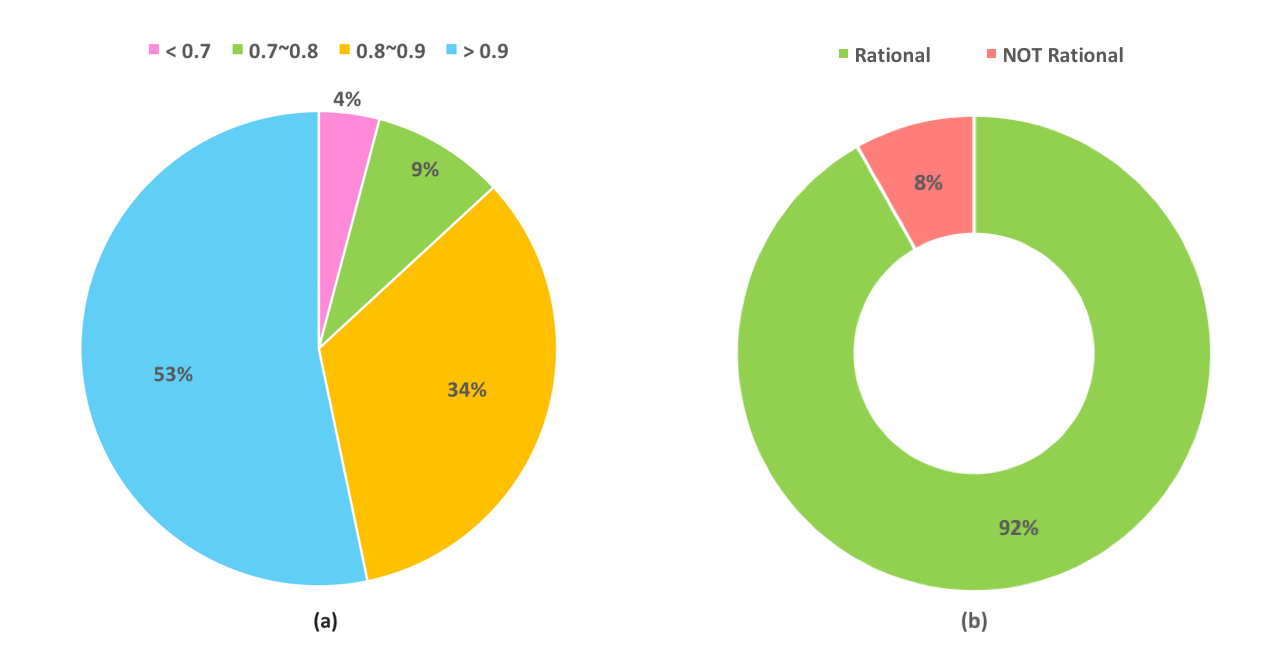}}
	\caption{Result of human evaluation of discovered visual themes on Corel5K dataset. (a): Result on accuracy of visual content description of visual themes. (b): Evaluators' responses on rationality of visual themes.}
	\label{fig:human}
\end{figure}

17 human subjects participated in the evaluation experiment and result is summarised in Fig. \ref{fig:human}. In (a) we can clearly see that more than half of discovered visual themes achieve an accuracy over 0.9 on visual content description, and only 4\% of them did not perform well on this task. In terms of rationality, 92\% of visual themes are voted as rational while the remaining 8\% are not. The experiment result demonstrates the effectiveness of discovered visual themes towards visual content description.

\section{Application of Visual Themes}
After human evaluation of visual themes, we further show potential applications of visual themes via three common computer vision experiments: example based image search, keyword based image search and image labelling. 

\subsection{Construct an Image Retrieval and Labelling Framework}
For building the framework, a vital issue need to be taken into consideration: how to design an effective data structure in terms of storage and speedy retrieval.  Inspired by \cite{fu2012fast}, we construct random forest using image features and discovered visual themes. In each random tree, we do binary split on visual features, and evaluate the split by computing histogram of visual themes. The well-know information gain\cite{chen2010efficient} is used as the objective function.

\begin{figure}[ht]
	\centering
	\centerline{\includegraphics[width=8cm,height=4cm]{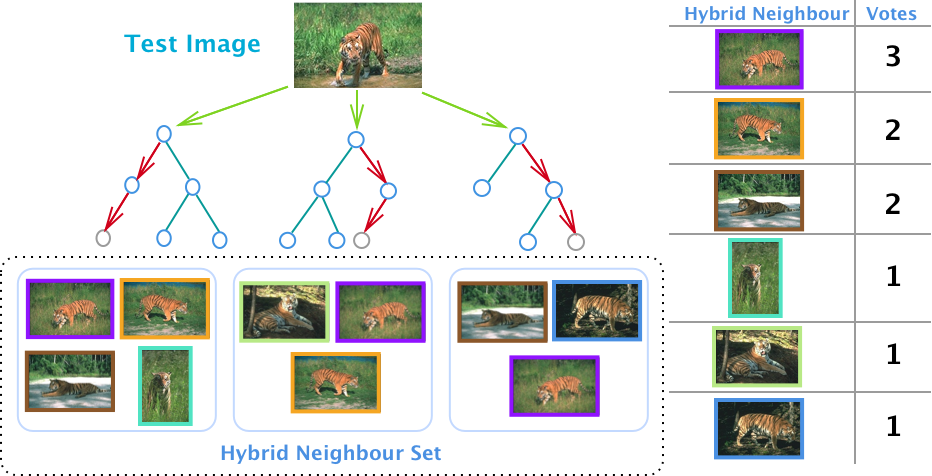}}
	\caption{Architecture of random forest for image retrieval and labelling. The visual feature of test image is put into the forest and similar images in training set will be found. Training images with higher frequency of occurrence will enjoy a higher rank in the retuned result.}
	\label{fig:parse}
\end{figure}

The architecture of random forest is illustrated in Fig. \ref{fig:parse}. Given a test image, we feed its visual feature into one random tree, and it keeps falling until it reaches a leaf node. Consequently, training examples under the same leaf node share similar or same visual themes with the test image. Here we name a related training example as a Hybrid Neighbour (HN). We successively feed the test image to all random trees and obtain the Hybrid Neighbour Set (HNS) which is formed by all HNs. Additionally, the frequency of occurrence for a single HN in HNS is defined as Hybrid Neighbour Vote (HNV). Apparently, a larger HNV indicates stronger similarity between a train image and the test image, and vice versa. 

\subsection{Example Based Image Search}
\textbf{Scenario}. The retrieval system accept an image as input and then returns a list of ranked images according to some similarity measure. In our case, we just put the test image into the random forest and obtain its HNS and corresponding HNVs. The returned images are then ranked by their HNVs following an descending order. Usually the top K results will be returned by the retrieval system.

\textbf{Data}. We work on the popular Corel5K \cite{duygulu2002object} benchmark which contains 4999 images. It is commonly split into 4500 image for training and the remaining 499 for testing, and 260 tags appear in both of these two sets. 

\textbf{Evaluation metric}. Since Corel5K dataset does not have ground truth images for this task, we use K-Nearest Semantic Measure (KNSM) defined in \cite{fu2012fast} as evaluation metric:

\begin{equation}
KNSM = \sum_{q=1}^{Q}\sum_{t=1}^{T} \sum_{k=1}^{K} \delta (H_{qk}, t)
\label{equ:knsm}
\end{equation}
where $Q$ is number of queries, $T$ denotes the number of tags contained in query image and $K$ represents that top K retrieved Hybrid Neighbours. $\delta (H_{qk}, t) = 1$ if query image $q$'s tag $t$ appears in its $k^{th}$ HN, and $\delta (H_{qk}, t) = 0$ if not. Hence a larger KNSM indicates stronger similarity between query image and its HNs since they share more tags.

\textbf{Parameter setting}. We eliminate tags whose visual content description levels (VCDLs) fall below 1.5, which results in 25 tags removed from original tag set. Next, 100 visual themes are obtained by clustering 235 remaining tags. Then we construct 400 random trees for image to image search. We also reproduce the result in \cite{fu2012fast} to justify the superiority of VSCC over tags. In terms of the baseline method, we select Joint Equal Contribution (JEC) \cite{makadia2010baselines} where various types of features are equally weighted for visual distance measurement, and is shown to perform well in image retrieval and annotation. 

\begin{figure}[ht]
	\centering
	\centerline{\includegraphics[width=10cm,height=4cm]{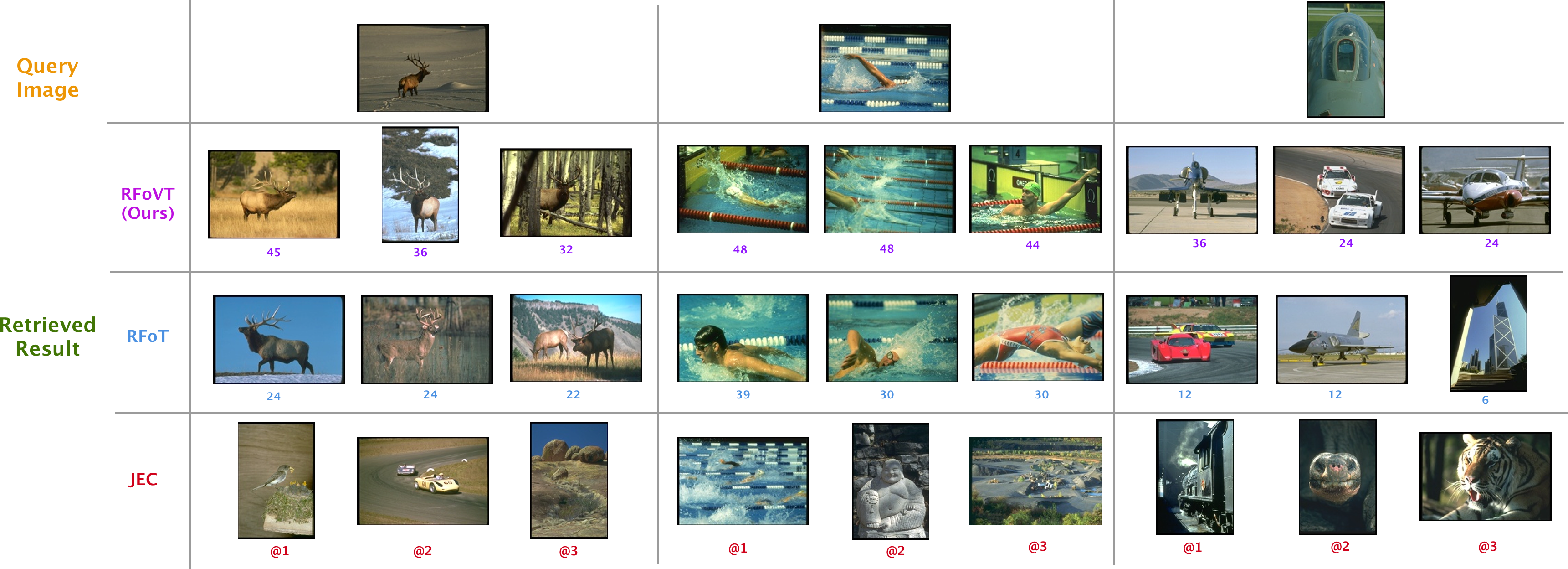}}
	\caption{Qualitative result of example based image search}
	\label{fig:iis}
\end{figure}

\textbf{Result}. Fig. \ref{fig:iis} shows some qualitative results of three methods: random forest on visual themes (RFoVT), random forest on tags (RFoT) \cite{fu2012fast} and JEC. Pink numbers under the result images denote their corresponding HNVs, the blue number is similar to HNV, but it's computed based on tags in stead of visual themes. Magenta numbers means the rankings of returned images using JEC method. Apparently RFoVT and RFoT greatly outperforms the JEC counterpart. Moreover, our RFoVT performs slight better than RFoT both in normal case (see first example) and hard cases (see last example).

\begin{figure}[ht]
	\centering
	\centerline{\includegraphics[width=8cm,height=5.3cm]{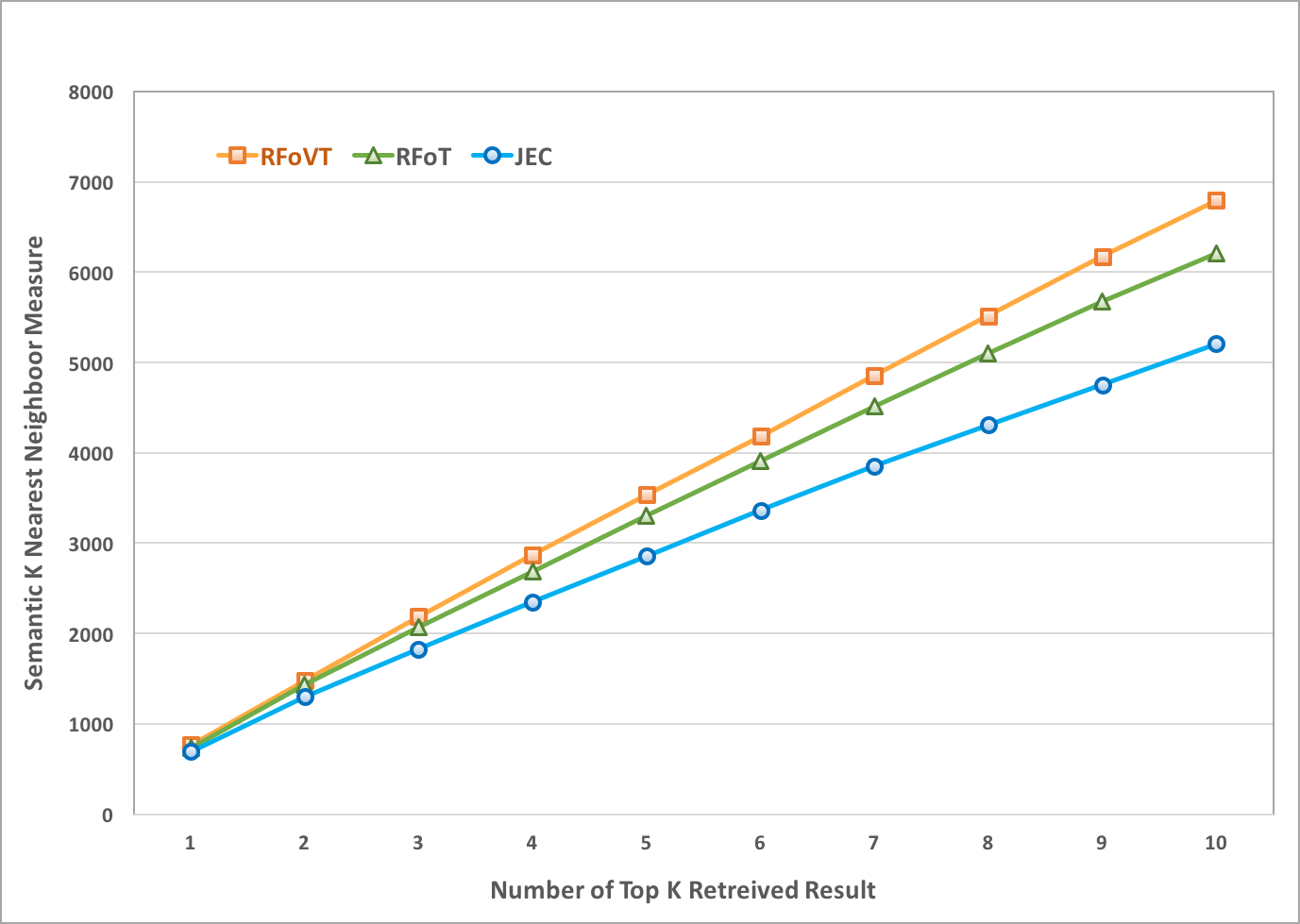}}
	\caption{KNSM measure of example based image search.}
	\label{fig:result}
\end{figure}

We also provide quantitative analysis using KNSM. We perform retrieval using all 499 testing images and result is illustrated in Fig. \ref{fig:result}. Clearly our method finds images with higher semantic similarity than the other two methods. Our success on this task demonstrates that visual themes are better than tags in terms of visual content description.

\subsection{Keyword Based Image Search}
\textbf{Scenario}. Given a query keyword, the retrieval system returns a collection of images that are most likely to contain that word. On this task, we tend to use a large image repository where training instances are annotated with tags while testing instances are not.

\textbf{Data}. We consider NUS-Wide-Lite dataset which contains 55,615 images, half of them (27,807) are used for training and the other half (27,808) for testing. We directly use 1,000 tags provided by the author for visual theme discovery. There are also 81 manually defined concepts available in dataset, each concept is represented with a single word. 

\textbf{Parameter setting and evaluation metric}. We first remove tags whose VCDLs are below than 2.5, then cluster 904 remaining tags into 300 visual themes. We build 400 random trees and evaluate the proximity between a test instance $T$ and a visual theme $c$ as:

\begin{equation}
p(T, c) = \frac{\sum_{n=1}^{N} \delta (h_{n}, c) v_{n}}{\sum_{n=1}^{N} v_{n} }
\label{equ:keyword} 
\end{equation}
where $N$ is the size of hybrid neighbour set (HNV) of instance $T$, $h_{n}$ denotes a hybrid neighbour (HN) in HNV, and $v_{n}$ denotes the hybrid neighbour votes (HNV) of $h_{n}$. $\delta (h_{n}, c)$ is an indicator function which equals to 1 if visual theme $c$ exists in $h_{n}$, and is equal to 0 otherwise. 

In experiment, we treat each visual theme as a whole keyword, that means searching using any tags within same visual theme will obtain same results. We compare the Mean Average Precision (MAP) achieved on visual themes (RFoVT) with five previous methods on 81 manually defined concepts, namely, K Nearest Neighbour (KNN), Support Vector Machine (SVM) \cite{witten2005data}, Entropic Graph Semi-Supervised Classification (EGSSC) \cite{subramanya2009entropic},  Label Exclusive Linear Representation (LELR) \cite{chen2011multi}, and Feature Analysis and Multi-Modality Fusion (CFA-MMF) \cite{ha2013correlation}. Additionally we repeat the work in \cite{fu2012fast} and construct another random forest using 81 manually defined concepts (RFoMC), then perform the same task. 

\begin{figure}[ht]
	\centering
	\centerline{\includegraphics[width=8cm,height=5.2cm]{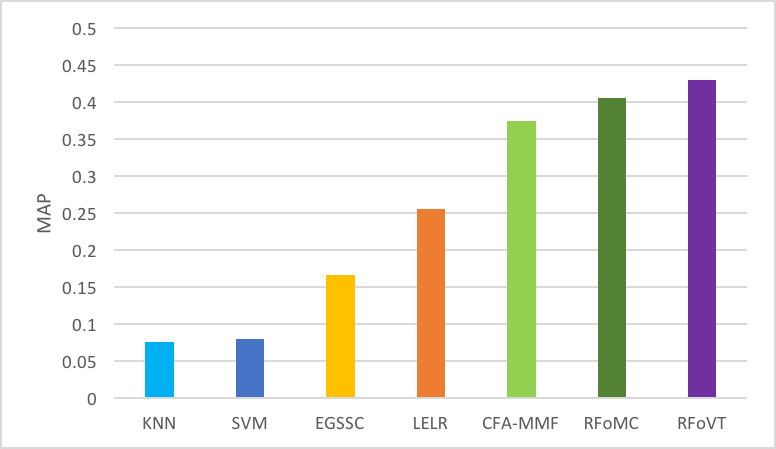}}
	\caption{MAP of keyword based image search on NUS-WIDE-Lite.}
	\label{fig:cis}
\end{figure}

\textbf{Result}. The overall results are shown in Fig. \ref{fig:cis}. We can clearly see that some of previous methods have achieved much higher MAP than the KNN baseline on 81 manually selected concepts, but they still fail to achieve a MAP over 40\%. While our random forest on visual themes (RFoVT) could obtain a MAP of 42.96\%. This result demonstrates automatically discovered visual themes could do better than manually selected concepts in terms of visual content representation.

\subsection{Image Labelling}
In order to further explore the potency of visual themes, we perform image labelling experiment on three well-known benchmarks: Corel5K \cite{duygulu2002object}, IAPR-TC12 \cite{escalante2010the} and a subset of ESP-game \cite{guillaumin2009tagprop:}. Table \ref{tab:il} provides details of three datasets and empirical settings of this task. 

\begin{table}[ht]
	\caption{Details of three image datasets and experimental parameters.}
	\begin{center}
		\begin{tabular}{l|c|c|c}  
			\hline
			Dataset & Corel5K & IAPR-TC12 & ESP Game \\ \hline  
			Number of training samples & 4500 & 17665 & 18689 \\         
			Number of testing samples & 499 & 1962 & 2081 \\        
			Number of tags & 260 & 291 & 268 \\ \hline
			$\alpha$ (Merging similarity matrix) & 0.15 & 0.3 & 0.2 \\
			Number of random trees & 400 & 400 & 400 \\
			Number of top voted HNs & 3 & 3 & 3 \\
			Number of tags returned & up to 5 & up to 5 & up to 5 \\ \hline
		\end{tabular}
	\end{center}
	\label{tab:il}
\end{table}

In this task, we do not perform tag filtering and only calculate the VCDLs for all tags. Given a test image, we put it into the random forest and obtain its HNs, and retain top voted $m$ HNs according to their HNVs. Then we collect all tags within these HNs and keep at most $n$ tags with highest VCDLs as final results. It's a natural approach since selected tags are visually and semantically connected to the test image. 

\begin{table}[ht]
	\caption{Image annotation results on three datasets.}
	\begin{center}
		\begin{tabular}{c|cc|cc|cc}
			\hline
			Dataset & \multicolumn{2}{c|}{Corel5K} & \multicolumn{2}{c|}{IAPR-TC12} & \multicolumn{2}{c|}{ESP Game} \\ \hline
			Method & Precision & Recall & Precision & Recall & Precision & Recall \\ \hline
			MBRM \cite{feng2004multiple} & 0.24 & 0.25 & 0.24 & 0.23 & 0.18 & 0.19 \\
			JEC \cite{makadia2010baselines}  & 0.27 & 0.32 & 0.28 & 0.29 & 0.22 & 0.25 \\
			TagProp \cite{guillaumin2009tagprop:} & 0.33 & 0.42 & 0.46 & 0.35 & 0.39 & 0.27 \\
			GS \cite{zhang2010automatic} & 0.30 & 0.33 & 0.32 & 0.29 & - & - \\
			SML+RF \cite{fukui2011multi-class} & 0.36 & 0.33 & 0.27 & 0.30 & - & - \\
			RF\_optimize \cite{fu2012random} & 0.29 & 0.40 & 0.45 & 0.31 & 0.41 & 0.26 \\
			RFoVT & \textbf{0.40} & \textbf{0.35} & 0.31 & 0.23 & 0.29 & 0.20 \\ \hline
		\end{tabular}
	\end{center}
	\label{tab:ilr}
\end{table}

We report average precision and average recall of image labelling with comparison to previous works in Table \ref{tab:ilr}. From the table we can see that our method (RFoVT) outperforms all previous methods on Corel5K dataset, but its performance falls behind TagProp \cite{guillaumin2009tagprop:} and RF\_optimize \cite{fu2012random} on the other two datasets. However, the success of TagProp largely depends on its tedious optimisation for each image and tag, which hinders its extension to large scale dataset. While RF\_optimize treats each tag as an independent unit and ignore their visual and semantic connection, which makes it less competent in dealing with noisy data. Note that web images in real world often come with considerable amount of redundant and unnecessary information. On the contrary, our image labelling method can be easily extended to large scale dataset, and can easily eliminate the majority of noisy dataset by applying tag filtering procedure. Although RFoVT does not perform very well on all datasets, it is quite simple yet efficient considering the intrinsic architecture of random forest. The result may be improved by adopting more sophisticated tag selection algorithm.

\section{Concluding Remarks}
In this paper, we put forward an unsupervised framework to automatically discover visual theme which can effectively describe visual contents. Then we perform manual evaluation to evaluate the quality of discovered visual themes, and show their potential applications in computer vision field via three common tasks. The results of example based image search, keyword based image search and image labelling experiments demonstrate the effectiveness and versatility of discovered visual themes.

\bibliographystyle{splncs}
\bibliography{refs}

\begin{thebibliography}{10}

\bibitem{duygulu2002object}
Duygulu, P., Barnard, K., de~Freitas, J.F., Forsyth, D.A.:
\newblock Object recognition as machine translation: Learning a lexicon for a
  fixed image vocabulary.
\newblock In: Computer Vision—ECCV 2002, Springer (2002)  97--112

\bibitem{chua2009nus}
Chua, T.S., Tang, J., Hong, R., Li, H., Luo, Z., Zheng, Y.:
\newblock Nus-wide: a real-world web image database from national university of
  singapore.
\newblock In: Proceedings of the ACM international conference on image and
  video retrieval, ACM (2009) ~48

\bibitem{escalante2010the}
Hugo Jair~Escalante, Carlos A~Hernandez, J.A.G.A.L.M.M.E.F.M.L.E.S.L.V.M.G.:
\newblock The segmented and annotated iapr tc-12 benchmark.
\newblock Computer Vision and Image Understanding \textbf{114} (2010)

\bibitem{guillaumin2009tagprop:}
Matthieu~Guillaumin, Thomas~Mensink, J.V.C.S.:
\newblock Tagprop: Discriminative metric learning in nearest neighbor models
  for image auto-annotation.
\newblock (2009)

\bibitem{sun2015automatic}
Sun, C., Gan, C., Nevatia, R.:
\newblock Automatic concept discovery from parallel text and visual corpora.
\newblock In: Proceedings of the IEEE International Conference on Computer
  Vision. (2015)  2596--2604

\bibitem{sadanand2012action}
Sadanand, S., Corso, J.J.:
\newblock Action bank: A high-level representation of activity in video.
\newblock In: Computer Vision and Pattern Recognition (CVPR), 2012 IEEE
  Conference on, IEEE (2012)  1234--1241

\bibitem{zhou2015conceptlearner}
Zhou, B., Jagadeesh, V., Piramuthu, R.:
\newblock Conceptlearner: Discovering visual concepts from weakly labeled image
  collections.
\newblock In: Proceedings of the IEEE Conference on Computer Vision and Pattern
  Recognition. (2015)  1492--1500

\bibitem{fang2015captions}
Fang, H., Gupta, S., Iandola, F., Srivastava, R.K., Deng, L., Doll{\'a}r, P.,
  Gao, J., He, X., Mitchell, M., Platt, J.C.,  et~al.:
\newblock From captions to visual concepts and back.
\newblock In: Proceedings of the IEEE Conference on Computer Vision and Pattern
  Recognition. (2015)  1473--1482

\bibitem{singh2015selecting}
Singh, B., Han, X., Wu, Z., Morariu, V.I., Davis, L.S.:
\newblock Selecting relevant web trained concepts for automated event
  retrieval.
\newblock In: Proceedings of the IEEE International Conference on Computer
  Vision. (2015)  4561--4569

\bibitem{deng2009imagenet}
Deng, J., Dong, W., Socher, R., Li, L.J., Li, K., Fei-Fei, L.:
\newblock Imagenet: A large-scale hierarchical image database.
\newblock In: Computer Vision and Pattern Recognition, 2009. CVPR 2009. IEEE
  Conference on, IEEE (2009)  248--255

\bibitem{miller1995wordnet}
Miller, G.A.:
\newblock Wordnet: a lexical database for english.
\newblock Communications of the ACM \textbf{38} (1995)  39--41

\bibitem{sadeghi2015viske}
Sadeghi, F., Divvala, S.K., Farhadi, A.:
\newblock Viske: Visual knowledge extraction and question answering by visual
  verification of relation phrases.
\newblock In: Computer Vision and Pattern Recognition (CVPR), 2015 IEEE
  Conference on, IEEE (2015)  1456--1464

\bibitem{divvala2014learning}
Divvala, S., Farhadi, A., Guestrin, C.:
\newblock Learning everything about anything: Webly-supervised visual concept
  learning.
\newblock In: Proceedings of the IEEE Conference on Computer Vision and Pattern
  Recognition. (2014)  3270--3277

\bibitem{babenko2014neural}
Babenko, A., Slesarev, A., Chigorin, A., Lempitsky, V.:
\newblock Neural codes for image retrieval.
\newblock In: Computer Vision--ECCV 2014.
\newblock Springer (2014)  584--599

\bibitem{shonkwiler1991computing}
Shonkwiler, R.:
\newblock Computing the hausdorff set distance in linear time for any lp point
  distance.
\newblock Information Processing Letters \textbf{38} (1991)  201--207

\bibitem{dubuisson1994modified}
Dubuisson, M.P., Jain, A.K.:
\newblock A modified hausdorff distance for object matching.
\newblock In: Pattern Recognition, 1994. Vol. 1-Conference A: Computer Vision
  \&amp; Image Processing., Proceedings of the 12th IAPR International
  Conference on. Volume~1., IEEE (1994)  566--568

\bibitem{mikolov2013distributed}
Mikolov, T., Sutskever, I., Chen, K., Corrado, G.S., Dean, J.:
\newblock Distributed representations of words and phrases and their
  compositionality.
\newblock In: Advances in neural information processing systems. (2013)
  3111--3119

\bibitem{bengio2006neural}
Bengio, Y., Schwenk, H., Sen{\'e}cal, J.S., Morin, F., Gauvain, J.L.:
\newblock Neural probabilistic language models.
\newblock In: Innovations in Machine Learning.
\newblock Springer (2006)  137--186

\bibitem{mikolov2014word2vec}
Mikolov, T., Chen, K., Corrado, G., Dean, J.:
\newblock word2vec (2014)

\bibitem{von2007tutorial}
Von~Luxburg, U.:
\newblock A tutorial on spectral clustering.
\newblock Statistics and computing \textbf{17} (2007)  395--416

\bibitem{simonyan2014very}
Simonyan, K., Zisserman, A.:
\newblock Very deep convolutional networks for large-scale image recognition.
\newblock arXiv preprint arXiv:1409.1556 (2014)

\bibitem{fu2012fast}
Fu, H., Qiu, G.:
\newblock Fast semantic image retrieval based on random forest.
\newblock In: Proceedings of the 20th ACM international conference on
  Multimedia, ACM (2012)  909--912

\bibitem{chen2010efficient}
Chen, X., Mu, Y., Yan, S., Chua, T.S.:
\newblock Efficient large-scale image annotation by probabilistic collaborative
  multi-label propagation.
\newblock In: Proceedings of the international conference on Multimedia, ACM
  (2010)  35--44

\bibitem{makadia2010baselines}
Makadia, A., Pavlovic, V., Kumar, S.:
\newblock Baselines for image annotation.
\newblock International Journal of Computer Vision \textbf{90} (2010)  88--105

\bibitem{witten2005data}
Witten, I.H., Frank, E.:
\newblock Data Mining: Practical machine learning tools and techniques.
\newblock Morgan Kaufmann (2005)

\bibitem{subramanya2009entropic}
Subramanya, A., Bilmes, J.A.:
\newblock Entropic graph regularization in non-parametric semi-supervised
  classification.
\newblock In: Advances in Neural Information Processing Systems. (2009)
  1803--1811

\bibitem{chen2011multi}
Chen, X., Yuan, X.T., Chen, Q., Yan, S., Chua, T.S.:
\newblock Multi-label visual classification with label exclusive context.
\newblock In: Computer Vision (ICCV), 2011 IEEE International Conference on,
  IEEE (2011)  834--841

\bibitem{ha2013correlation}
Ha, H.Y., Yang, Y., Fleites, F.C., Chen, S.C.:
\newblock Correlation-based feature analysis and multi-modality fusion
  framework for multimedia semantic retrieval.
\newblock In: Multimedia and Expo (ICME), 2013 IEEE International Conference
  on, IEEE (2013)  1--6

\bibitem{feng2004multiple}
S~L~Feng, R~Manmatha, V.L.:
\newblock Multiple bernoulli relevance models for image and video annotation.
\newblock (2004)

\bibitem{zhang2010automatic}
Shaoting~Zhang, Junzhou~Huang, Y.H.Y.Y.H.L.D.M.:
\newblock Automatic image annotation using group sparsity.
\newblock (2010)

\bibitem{fukui2011multi-class}
Motofumi~Fukui, Noriji~Kato, W.Q.F.X.:
\newblock Multi-class labeling improved by random forest for automatic image
  annotation.
\newblock (2011)

\bibitem{fu2012random}
Hao~Fu, Qian~Zhang, G.Q.:
\newblock Random forest for image annotation.
\newblock European Conference on Computer Vision (2012)

\end{thebibliography}



\end{document}